\documentclass[preprint,12pt]{elsarticle}

\usepackage[utf8]{inputenc}
\usepackage[T1]{fontenc}
\usepackage{amssymb}
\usepackage{amsmath}
\usepackage{booktabs}
\usepackage{tabularx}
\usepackage{array}
\usepackage{hyperref}
\usepackage{verbatim} 
\usepackage{graphicx}

\journal{Journal of Biomedical Informatics}

\begin{document}

\begin{frontmatter}

\title{Do It Right! A Methodology for Successful NLP System Development}

\author[1]{Patterson OV\corref{cor1}}
\ead{ovpatterson@gmail.com}
\author[1]{South B}
\author[1]{Workman TL}
\author[1]{DuVall SL}

\cortext[cor1]{Corresponding author}

\affiliation[1]{organization={[Affiliation placeholder]},
               addressline={},
               city={},
               postcode={},
               state={},
               country={}}

\begin{abstract}
Natural language processing (NLP) is a common method for supplying data to
clinical research and decision making by extracting information from electronic
medical records \cite{Neveol2015,Neveol2016,Friedman2006a}. Numerous textbooks
and tutorials describe specific algorithms and applications for text processing
\cite{Jones1994,Sager1987,Doan2014a,Nadkarni2011,Grishman1997,Schuemie2005,Sarawagi2008a,Meystre2008},
yet algorithmic knowledge is only one ingredient of a successful NLP project.
Drawing on the available literature, this paper presents a stepwise approach
that applies the Systems Development Life Cycle (SDLC) to projects that rely on
data extraction through language processing.
\end{abstract}


\end{frontmatter}

\section{Introduction}

NLP encompasses a broad range of algorithms for the computerized processing of
unstructured text. For roughly its first 40 years after emerging as a discipline
in the 1950s, NLP was largely a promise of the future: linguists studied
language structure, computer scientists developed algorithms, and hardware
engineers expanded computational capacity. Many systems were built, but few
crossed from academic research into practical use
\cite{Neveol2016,Jones1994,Sager1987,Demner2009}. Since the mid-1980s, the
growth of electronic data, greater computing power, and an expanding open-source
community have lowered barriers to entry and broadened NLP's use, including
clinical text processing for health outcomes research \cite{Spyns1996}.

NLP has long since moved beyond computer science to become central to clinical
informatics and biomedical research \cite{Nadkarni2011}. As interest has grown,
experts have produced a wealth of learning materials covering individual tasks,
including parsing, part-of-speech tagging, and semantic role labeling, as well
as machine learning and system architecture
\cite{Jones1994,Sager1987,Doan2014a,Nadkarni2011,Grishman1997,Schuemie2005,Sarawagi2008a,Meystre2008}.
This abundance of resources and freely available implementations can suggest
that building an information extraction system for a given use case requires
little specialized effort.

The recent proliferation of large language models (LLMs) has renewed this
impression. Models capable of answering clinical questions, summarizing notes,
and extracting structured data are now accessible to any researcher with an
application programming interface (API) key. The apparent ease suggests that
reliable extraction is now within routine reach. That picture is an illusion.
These models fabricate plausible content not present in the source, omit
information that is there, and produce inconsistent results from prompts that
say the same thing differently. These models are impressive but do not remove
the need for the underlying process. Instead, they make that need less visible.

NLP development carries the same project risks as any other software
undertaking. The clinical literature rarely reports failures, but the Management
Information Systems literature documents them extensively \cite{Kappelman2006},
and clinical NLP has no particular immunity. Business practitioners address
these risks through the Systems Development Life Cycle (SDLC), a formal
sequence of steps for developing computerized solutions. Applying the SDLC
deliberately gives clinical researchers a structured means of reducing the
likelihood of failure.

Applying clinical NLP also resembles retrospective manual chart abstraction.
The primary difference is the use of computerized algorithms in place of human
reviewers. The extensive chart-abstraction literature, therefore, offers
relevant lessons in project success and failure
\cite{Vassar2013,Gearing2006,Knake2016}. The recent arrival of large language
models draws this parallel even closer. Where earlier rule- and feature-based
systems matched patterns that bore little resemblance to human reading, LLMs
interpret narrative text in a way that approximates how a human abstractor reads
a chart. Like a human abstractor, they can misread, infer beyond the evidence,
or apply an interpretation inconsistently. The similarity is therefore no longer
merely conceptual: the same disciplined practices that make manual chart
abstraction reliable---such as explicit definitions, reviewer training,
adjudication, and measured agreement---apply directly to LLM-based extraction,
and the same process discipline is required to keep either from failing.

This paper introduces the SDLC steps and offers specific recommendations for
carrying them out when building a clinical NLP information extraction system.

\section{Systems Development Life Cycle}

Tutorials, review articles, and textbooks tend to emphasize algorithms while
neglecting the fact that application development is a process that must be
managed. Publications on failed projects are rare in clinical informatics, but
the Management Information Systems literature reports many, and mismanagement is
the most frequently cited cause of failure \cite{Kappelman2006}.

The SDLC has been used since the 1960s, when it was formalized to manage
complex software projects for large-scale business systems. Although many
variants exist for different project types, all share the same core phases:

\begin{enumerate}
  \item Planning
  \item Analysis
  \item Design
  \item Implementation
  \item Testing
  \item Deployment
  \item Maintenance
\end{enumerate}

The sections below discuss each phase and show how to carry an information
extraction project from start to finish. Skipping a required step can
jeopardize the entire project or, at minimum, waste time and effort. Our focus
is on clinical applications of NLP, and on information extraction specifically.
Tasks that involve text generation, such as summarization, introduce different
design and evaluation requirements and are outside the scope of this
methodology. Even a small research project that applies NLP to a single dataset
benefits from being managed properly.

\section{Planning}

Planning comes first. The project team must identify the intended purpose of
the future system, since different goals impose different requirements and
constraints, and then assess whether that goal is attainable through a
feasibility analysis.

\subsection{Project Purpose}

The system's purpose should be conceptualized during planning. Project sponsors
typically begin with a genuine organizational need \cite{Yeoh2010}, and the
purpose must deliver real value to the organization it serves \cite{Asosheh2010}.
Baccarini defines project purpose as ``the intended near-term effects on the
user of the product as a result of utilizing the project's outputs,'' measured
by how well the system meets user needs \cite{Baccarini1999}. Purpose is
intertwined with the project's overall goal and provides the means of achieving
it \cite{Baccarini1999}. A project should have a single central purpose so that
team effort stays focused \cite{Davis2005}. That purpose can be framed as
requirements paired with acceptable limitations. Consider three representative
scenarios (Table~\ref{tab:purpose}).

\begin{table}[htbp]
\centering
\caption{Representative project purpose scenarios.}
\label{tab:purpose}
\begin{tabularx}{\linewidth}{>{\bfseries}p{3.5cm} p{4.5cm} p{4.5cm}}
\toprule
Purpose & Primary Requirement & Acceptable Limitations \\
\midrule
Novel algorithm or proof-of-concept &
  Original codebase and algorithm suitable for publication &
  Slow performance, high error rate, small document set \\[6pt]
Applied extraction for clinical research &
  Highly accurate dataset of specific data elements &
  No originality required; code reuse preferred \\[6pt]
Commercial or enterprise system &
  Robust, scalable, flexible, with configurable input and output &
  Limited scope per deployment \\
\bottomrule
\end{tabularx}
\end{table}

\begin{enumerate}
  \item \textbf{A novel algorithm or proof-of-concept system.} The requirement
  is an original codebase and algorithm suitable for publication. Depending on
  the research question, slow performance, a high error rate, or a small
  document set may be acceptable. Like fundamental research, such systems may
  have no commercial use case but can advance general understanding of language
  processing.

  \item \textbf{Applied extraction for a clinical research study.} The
  requirement is a highly accurate dataset of specific data elements. Here,
  originality is undesirable and code reuse is preferred, because the system is
  only a means to obtain the data. Reusing accepted approaches and existing
  modules yields a custom solution with the least effort, and high accuracy is
  often achieved by narrowing the clinical subdomain.

  \item \textbf{A commercial or enterprise system.} These must be robust,
  scalable, and flexible enough to process large, diverse document sets without
  system-wide failure and to accept configurable input and output formats. Such
  qualities are achieved by limiting the system's scope.
\end{enumerate}

In every case, the system should address the specific needs recorded in the
research proposal or planning documentation, and success criteria must be
defined, including how to measure the extent to which the system meets its
goals.

\subsection{Project Scope}

Inadequate requirements and undefined scope are recognized causes of project
failure \cite{Kappelman2006,Vassar2013,Nelson2007}. Identifying target variables
and operationalizing their definitions logically and reproducibly is essential
to any study. Clinical documents contain vast amounts of information about a
patient and their environment, expressed as concepts, values, and relationships.
A single concept can be expressed by many different words, phrases, and symbols.

The difficulty of a study depends on the complexity of the concept of interest.
Simple concepts have clear, widely agreed definitions. Patient weight, for
example, is a routinely measured vital sign expressed in pounds or kilograms.
Complex variables either lack a generally accepted definition, such as
``culture'' \cite{Iwamoto2010}, or combine multiple simple variables, such as
``congestive heart failure'' \cite{Meystre2016}, evidence of homelessness
\cite{Gundlapalli2013}, or other social determinants of health.

Because projects often run for months or years, formal documentation of variable
definitions is a required part of planning. A \textbf{concept sheet} records
the detailed definition and prevents unplanned drift. Beyond decomposing complex
concepts into simple ones, concept sheets specify expected ranges and units for
numeric variables and allowable values for categorical variables.

Concept sheets are also a key communication device. NLP-supported clinical
studies are conducted by multidisciplinary teams combining clinical and
technical expertise \cite{Neveol2015,Neveol2016}, so the sheets must be shared
among all personnel. Their level of formality can vary with project size, but
the definitions must be understood and agreed upon by the whole team and remain
readily accessible throughout the project. When definitions change, the updates
must be distributed immediately.

Failures typically stem from problems with people, processes, or product risks
\cite{Kappelman2006}, and in NLP work people and processes are tightly linked.
Kappelman et al.\ list ``no change control process'' among the top failure
risks, an issue that a routinely reviewed concept sheet helps manage, and cite
missing risk-analysis documentation as another. Poor estimation of risk is the
single largest source of failure \cite{Nelson2007}, making risk analysis a
crucial part of any feasibility assessment.

\section{Analysis}

Once goals are defined, the team must estimate the feasibility of achieving them
and, if feasible, define the target corpus.

\subsection{Feasibility Analysis}

Feasibility hinges on two questions, applicable regardless of the extraction
approach: (1)~Does the information exist in the available clinical text?
(2)~Can it be extracted with sufficient accuracy?

Answering the first question requires manual inspection of the target corpus.
Retrospective clinical studies rely on data collected for care rather than
research, so the variables of interest may or may not be routinely documented.
Clinicians in the target subdomain can usually describe how information is
recorded, though practices vary across sites and change over time. When the
extraction approach uses a large language model, this initial characterization
requires particular care. Hallucination---the generation of plausible but
unsupported content---means that apparent extraction success is not evidence
that the information exists in the text.

A lightweight manual chart review of a representative sample can confirm whether
target concepts are present. When they are not consistently documented, concept
sheets may need to define alternative or surrogate variables. Chest radiograph
reports, for instance, rarely state a pneumonia diagnosis outright
\cite{Liu2013}, so identifying pneumonia may instead rely on mentions such as
``opacity'' and ``consolidation'', even though ``opacity'' is not the diagnosis
itself.

The second question depends on how well the approach handles ambiguity. Two
challenges, semantic ambiguity and contextual ambiguity, largely determine
whether high-accuracy extraction is achievable by any method, and both are
discussed in the subsections below. For LLM-based approaches, published
benchmark performance is rarely a reliable guide to accuracy on a specific
clinical task. Feasibility must be assessed on representative examples from the
target corpus, not inferred from general capability.

\subsection{Semantic Ambiguity}

A \textbf{term} is a string in text that represents a relevant clinical entity.
``LVEF,'' for example, is a term for an echocardiogram measurement. A
\textbf{concept} is a standardized representation of that entity, linked to all
terms that can express it: ``ef,'' ``LVEF,'' ``lv ej frac,'' and ``ejection
fraction'' are distinct terms for the same concept, left ventricular ejection
fraction.

Semantic ambiguity is the number of concepts that a given term can map to.
Short abbreviations are especially ambiguous: ``PE'' can mean pulmonary edema,
peripheral edema, pulmonary embolism, physical education, or physical
examination, among others. Resolving such cases may require extensive
word-sense disambiguation. Fully spelled-out terms are far less ambiguous.
``Left ventricular ejection fraction,'' for example, has only one meaning. The
more possible concepts a term maps to, the harder disambiguation becomes and the
less feasible high performance is. In extreme cases, even a human reader cannot
reliably interpret a term without contextual clues.

The \textit{prevalence} of the competing meanings matters as much as their
number. ``CCS,'' for example, can mean Canadian Cardiovascular Society score or
cubic centimeters. Cubic centimeters appears constantly as a unit of measure,
whereas the severity score is stated very rarely, so a system seeking the score
must sift through millions of mentions to find the fewer than 0.1\% that are
relevant.

\subsection{Contextual Ambiguity}

Once a term is mapped to a concept, the context in which it appears must be
identified. Three contextual axes dominate the clinical literature: assertion
(also called negation), temporality, and experiencer. Assertion may be positive,
negative, or possible. Temporality may be historical, current, or hypothetical.
The experiencer may be the patient or someone else. Other axes, such as severity
or anatomic location, may be relevant to a particular study.

Some concepts use unambiguous terms but appear in highly variable contexts.
``Fever,'' for example, is semantically clear (core body temperature above
$98.6^\circ$F, setting aside phrases like ``yellow fever'') yet can occur in
many contexts (see Table~\ref{tab:fever}).

\begin{table}[htbp]
\centering
\caption{Contextual variation for the term ``fever.''}
\label{tab:fever}
\begin{tabularx}{\linewidth}{Xccc}
\toprule
Example mention & Assertion & Temporality & Experiencer \\
\midrule
``The patient has a fever.'' & Positive & Current & Patient \\
``No fever noted.''          & Negative & Current & Patient \\
``Possible low-grade fever.'' & Possible & Current & Patient \\
``History of fever.''        & Positive & Historical & Patient \\
``Mother had fever during pregnancy.'' & Positive & Historical & Other \\
``Watch for fever if symptoms worsen.'' & Possible & Hypothetical & Patient \\
\bottomrule
\end{tabularx}
\end{table}

Conversely, rare concepts that closely resemble others---such as specific HIV
gene mutations---are mentioned only when found present. In the passage ``A
genotype \ldots\ showed significant HIV drug resistance, including the following
mutations in HIV reverse transcriptase: M41L, L210W, and T215Y,'' the possible
mutations are too numerous to enumerate negatively, so a mention is always
affirmed, current, and about the patient.

Manual chart review is often used for feasibility analysis, confirming both
that target information is documented and that it is documented in a
machine-accessible way. Agreement among multiple abstractors on the same
documents provides an upper bound on achievable NLP performance.

Feasibility analysis may show that a fully automated system is impractical,
requiring the project definition to change. When target concepts are present
but highly ambiguous, automated extraction produces a high error rate.
Alternatives include fully manual or NLP-assisted manual review. Fully manual
extraction is accurate but slow, suitable only for small datasets. For large
datasets, a high-recall NLP system can preselect instances for manual
disambiguation. In our work on advanced basal cell carcinoma, a common cancer
that rarely reaches advanced stage, fully automatic extraction was insufficiently
accurate given the variability and low prevalence of relevant mentions, so we
used a simple NLP system to label all mentions and then curated them manually to
detect advanced stage. When target concepts are neither well documented nor
tractable at scale, structured-data surrogates may replace the NLP solution
altogether.

Once feasibility is established, analysis turns to document selection.

\subsection{Document Selection}

NLP uses narrative text as its primary source. The electronic medical record
(EMR) accumulates documents from every patient encounter, and patients with
complex histories may have thousands of progress notes, nursing notes, and lab
and pathology reports. Selecting relevant documents reduces both the volume to
be processed and the semantic variability of the corpus. Documents must also be
selected for system training and testing.

\textbf{Document selection for processing.} The most basic approach limits the
corpus to documents belonging to the study cohort and generated within the study
period. Filtering by document title is also common and works well where document
titles are strictly controlled. The Logical Observation Identifiers Names and
Codes (LOINC) document ontology was created to standardize this, classifying
documents along five axes: subject-matter domain, author role, clinical setting,
type of service, and kind of document. The Observational Medical Outcomes
Partnership (OMOP) common data model recommends standard document titles to
promote harmonization, and the Health Level 7 (HL7) standard requires LOINC.
In practice, however, many EMRs allow custom titles. The Veterans Affairs (VA)
VistA system, for example, permits custom document titles across more than 50,000
clinicians and 130 disconnected installations, so the number of titles keeps
growing. Title-based selection is viable where HL7 is followed but unreliable
where naming is not enforced.

When computational cost is high and relevant mentions are rare, additional
information-retrieval steps may be needed to pre-filter the corpus before
processing. This consideration is especially important for LLM-based approaches,
where most providers charge per token and costs scale directly with corpus size.
Passing large volumes of irrelevant clinical text through an LLM is expensive
and unnecessary. A lightweight pre-filtering step, such as keyword matching or
document classification, can substantially reduce the number of documents
processed and the cost.

\textbf{Document selection for annotation.} Careful selection reduces the
expense of manual review. Selecting a random sample is simplest but can be
suboptimal when target information is infrequent or the corpus is linguistically
diverse. Stratified random sampling is an alternative when language is expected
to vary across identifiable strata, such as clinical subdomain, locality, or
time, as linguistic drift introduces new terms. Because language is more
consistent within than across strata, sampling within each stratum captures a
broader range of examples, which is especially useful when strata differ in
size.

\textit{Sample size.} In NLP studies, much as in qualitative research,
training-set size is often determined by a maximum-variation approach: positive
cases are reviewed and annotated until no new concepts, vocabulary, or lexical
variants emerge \cite{Topaz2016}.

The examples in this paper focus on \textbf{supervised} NLP development, in
which training data is manually labeled by human reviewers, as opposed to
unsupervised approaches that learn patterns statistically. Many real systems
combine supervised, unsupervised, and semi-supervised methods. We use supervised
examples for simplicity. Supervised development typically requires a manually
reviewed \textbf{truth set} produced through annotation.

\section{Annotation}

NLP development requires adequate data for training and evaluation, and manual
annotation is a common way to produce it. Annotation is a schema-based manual
review that identifies spans of text representing target information classes and
may also assign attributes or identify relations between classes. It is, in
effect, human information extraction, and projects may require annotating
anywhere from small to very large document sets. Annotated corpora tell both
programmers and machines what to extract or classify.

Annotation is time-consuming, expensive, and error-prone, and demands
considerable reviewer effort, but human review of some kind is needed to
understand variable definitions, refine target classifications, and further
define scope. It also reveals the kinds of examples automated extraction will
require. These steps are typically iterative: identify examples, refine the
guidelines, and reassess reviewer agreement. Such examples are essential for
confirming that a system works as expected before it is tested on held-out or
previously unseen documents. Annotation can be cognitively demanding. Some tasks
require expert judgment while others can be done by non-experts.

For annotation to function as a valid reference standard, the data available to
annotators must match the data the NLP system will see. The most useful
annotation replicates the pipeline exactly: annotators work from the same source
documents, in the same form, and record output in the same schema the system
will produce. When annotators have access to additional context---such as
structured data fields, prior notes, or other documents outside the system's
scope---the reference standard reflects capabilities the system does not
possess. Measured performance against such a standard will be lower than
annotation agreement suggests is achievable, obscuring where the system actually
fails and making targeted improvement more difficult. Annotation design should
therefore begin with an explicit specification of what the NLP system will and
will not have access to.

\subsection{Annotation Guidelines}

Once extraction targets are identified, an annotation guideline is written with
the study investigators to specify the targets, define concepts and attributes,
and describe relationships between concepts. Good guidelines provide clear
examples of what to annotate (inclusions), how to classify it, and what to
leave out (exclusions).

For most tasks, individual mentions are the targets the system will identify
and extract. Mentions may carry one or more attributes describing contextual
features or state. The most common attribute types---negation, experiencer, and
temporality---describe context, and the guideline specifies exactly how each is
assigned. Tasks may also identify logical relations between mentions, such as a
disease and the drugs used to treat it. Guidelines may additionally cover use
of the annotation tools. The SDLC integrates naturally here, particularly in
developing and refining guidelines and running an annotation campaign.

\section{Design}

System design identifies the key elements of the future system along with their
inputs, outputs, and processing steps. A complex NLP problem becomes manageable
when broken into its most basic tasks. Just as matter reduces to a few kinds of
atoms, NLP problems reduce to two atomic tasks: \textbf{search} and
\textbf{classification}. A third task, summarization, is sometimes treated as
atomic alongside these two. Summarization, however, combines extraction with
generation---it identifies relevant content and produces new text to represent
it. Generation is a fundamentally different kind of task from extraction and is
outside the scope of this paper. Inefficiency arises when a designer treats a
set of distinct tasks as a single process.

Several design approaches are possible. In a well-documented domain, existing
resources such as dictionaries and ontologies can supply the logical units.
Data-driven approaches instead rely on manual annotations for examples. Ensemble
systems apply multiple methods in parallel.

An \textbf{information model} represents the concepts, relationships, rules,
constraints, and operations that define the data semantics of a domain. A
\textbf{knowledge base} encodes that model for a specific implementation.
Knowledge bases are commonly built either by having experts specify rules or by
learning a model from human-annotated text. Five methods for identifying and
interpreting concepts are in common use: rules, patterns, machine learning,
large language models, and hybrid systems that combine these approaches. When a
large language model is chosen, the prompt takes on the role of both the concept
sheet and the annotation guideline. It must define the target concept, specify
what counts as an instance, and describe how to handle ambiguous cases
\cite{Guo2024,Syrstad2024}. Retrieval-augmented generation, which grounds LLM
output in a curated external knowledge base, is the modern counterpart to the
information model \cite{RAG2024a,RAG2024b}. For named-entity recognition and
classification tasks, fine-tuned domain-specific encoder models continue to
outperform general-purpose LLMs, and the choice of method should be driven by
task requirements, not by the novelty of the technology \cite{Guo2024,Syrstad2024}.

Design also operates at different levels of analysis.

\begin{itemize}
  \item \textbf{Mention level.} The most common approach: human reviewers
  classify each mention to a concept, and that training data teaches the system
  to identify and classify spans of text. A document may contain many mentions,
  and they need not agree.

  \item \textbf{Document level.} The entire document is classified. A patient
  may have from a handful to thousands of documents.

  \item \textbf{Episode level.} At this level, mentions across multiple
  documents are aggregated to capture a clinical event at a specific date or
  over a defined window, such as an episodic flare in Crohn's disease or
  cholecystitis. A patient may have many events or none, and mentions need not
  agree. Disagreements are reconciled as part of the extraction logic.

  \item \textbf{Patient level.} All mentions and document classifications are
  combined into a single patient-level classification. Extraction results at
  every level are compared against a human-defined truth set.
\end{itemize}

Once the required tasks and goals are set, the full sequence of steps can be
designed. This sequence, called the \textbf{pipeline}, is what each document
passes through, with each step adding metadata about document structure or
content. A pipeline may include preprocessing to normalize formatting, remove
extraneous characters, or combine documents from different specialties, tables,
or EMR locations.

The accepted way to store annotations is as metadata: annotations are kept in a
separate structure linked to the source document by the start and end indices of
each labeled span. This leaves the original document unmodified and preserves
all data unambiguously. Annotation viewers read both files and combine them
visually for human interpretation. LLM-based systems complicate this convention.
Unlike rule- or pattern-based systems that identify explicit text spans, an LLM
arrives at a conclusion through internal attention mechanisms that are not
directly observable. It is therefore not always possible to pinpoint the exact
portion of text the model relied on, which makes error analysis more difficult
and reduces the interpretability of system output.

\section{Implementation}

Implementation is the focus of most systems-engineering courses, and the first
three SDLC phases are largely independent of it.

\begin{figure}[htbp]
\centering
\includegraphics[width=0.85\linewidth]{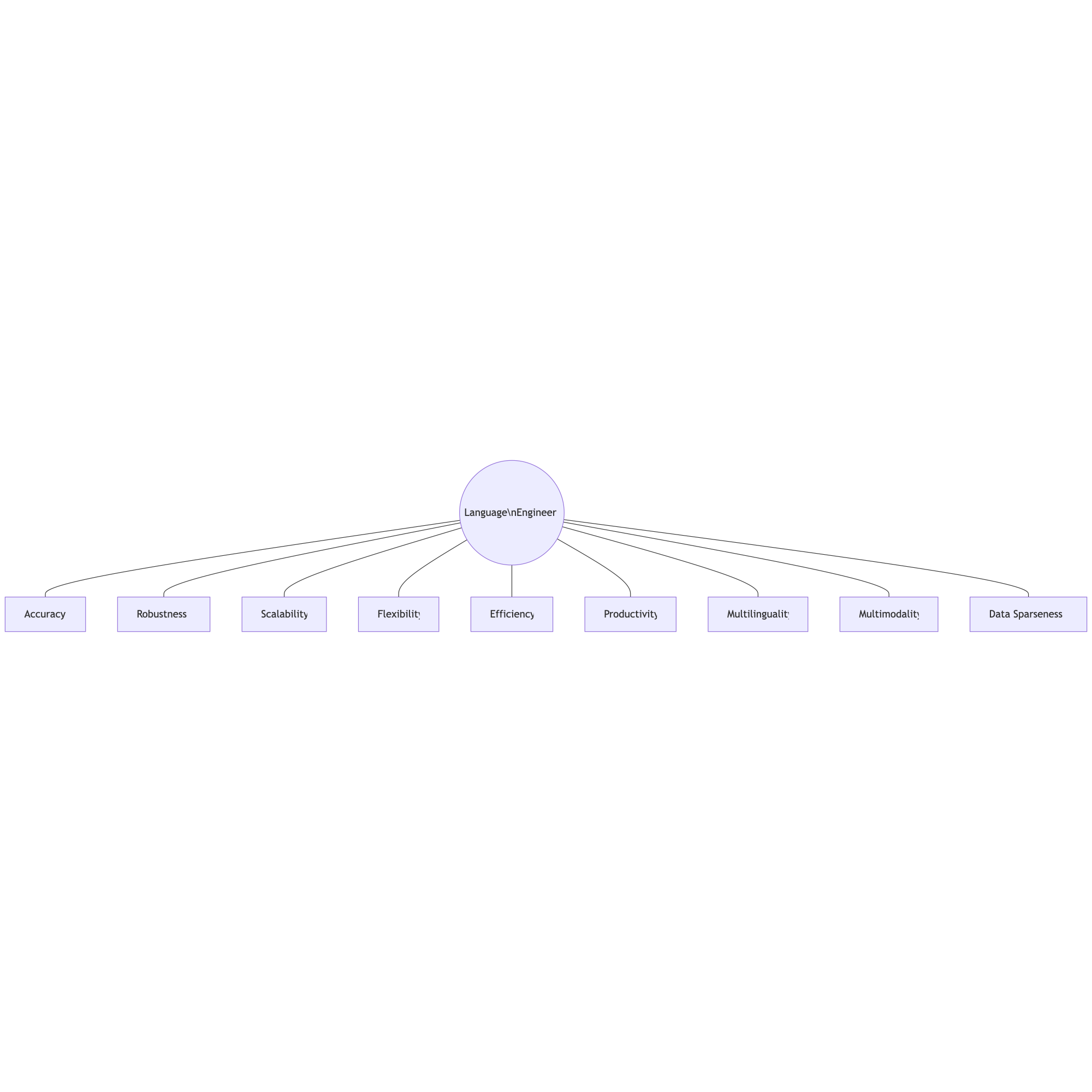}
\caption{Concerns of the language engineer (after Leidner, 2003).\protect\footnotemark}
\label{fig:leidner}
\end{figure}

Leidner (2003) summarized the key complexities of building an NLP system
particularly well. Several of his concerns are worth emphasizing
(Figure~\ref{fig:leidner}).

\footnotetext{Source: Leidner JL. Current issues in software engineering for
Natural Language Processing. Proceedings of the HLT-NAACL 2003 Workshop on
Software Engineering and Architecture of Language Technology Systems (SEALTS
'03). ACL, Morristown, NJ; 2003. pp.~45--50.}

\textbf{Accuracy.} Text processing is inherently difficult, and correct output
is never guaranteed. Potential errors must be accounted for whenever NLP output
is used.

\textbf{Flexibility.} Input and output formats should be configurable rather
than hard-coded. Hard-coded values limit utility and cause run-time failures
when an expected file is missing. For LLM-based systems, prompts carry the same
risk: a prompt embedded in code as a string literal cannot be updated without a
code change and is invisible to non-developers. Prompts should be stored as
configurable, version-controlled artifacts alongside the system's other
parameters.

\textbf{Scalability.} A system should process large volumes within a reasonable
time, often through distributed architecture that spreads pipeline instances or
stages across multiple network nodes.

\textbf{Reuse.} Despite many researchers and shared code repositories, reuse of
fully configured, comprehensive NLP pipelines is low. Different use cases call
for different pipeline designs, so end-to-end systems rarely transfer without
substantial modification. Only a few individual modules are broadly reused,
among them Hugging Face Transformers \cite{Wolf2020}, Stanford's Stanza
\cite{Qi2020}, and spaCy \cite{Honnibal2020}. Integration remains difficult
when developers use different languages, deprecated environments, or divergent
paradigms. Leidner lists several barriers to reuse:

\begin{itemize}
  \item Lack of awareness of existing components
  \item Lack of trust in component quality
  \item Mismatch between component properties and project requirements
  \item Licensing, cost, organizational, or technical barriers such as platform
  incompatibility, dependency conflicts, and poor documentation
\end{itemize}

For LLM-based systems, reusability faces an additional challenge: the pace of
new model releases is rapid, and performance varies substantially across models
and across versions of the same model. A pipeline validated against one model
version may behave differently after an update, and switching to a newer or
different model often requires re-evaluation rather than straightforward
substitution. This makes LLM components harder to treat as stable, reusable
building blocks.

Leidner describes this tension through a \textit{productivity pyramid} that
maps implementation approaches along a trade-off between short-term productivity
and long-term reusability. At the top of the pyramid are ready-made solutions
that reduce development time but offer limited control and create dependency on
external vendors. At the base are custom-built, modular components that require
more investment but yield systems that are more controllable, maintainable, and
adaptable over time. An increasing number of teams now operate toward the top of
the pyramid by building on NLP utilities provided by cloud platforms such as
Azure, AWS, Google Cloud, and Databricks. These services offer pre-built
functions for common tasks including entity recognition, sentiment analysis, and
document classification, reducing development effort at the cost of reduced
control over model behavior and dependency on vendor roadmaps and pricing.
Existing solutions are widely available but rarely do exactly what is needed.
Planning, analysis, and validation remain essential regardless of where a team
sits on the pyramid, and the best outcomes typically come from a flexible
framework built from compatible modules rather than a single off-the-shelf
system.

LLM-based systems introduce a distinct set of implementation decisions. Prompt
engineering is the primary tool for steering model behavior without modifying
model weights, and few-shot examples embedded in the prompt can substantially
improve performance on structured extraction tasks. When prompting alone is
insufficient, the model can be fine-tuned on domain-specific examples at a
fraction of the cost of full retraining. The choice between local deployment and
hosted APIs involves trade-offs in cost, latency, and data governance.
Open-source models such as Llama~3 and Mistral can now be run locally using
serving frameworks such as Ollama or vLLM, keeping clinical text entirely within
the organization's infrastructure and eliminating external data transmission.
For organizations preferring hosted services, the same major cloud providers
mentioned above also offer LLM-specific services, including Azure OpenAI, AWS
Bedrock, and Google Vertex AI, with Health Insurance Portability and
Accountability Act (HIPAA)-eligible configurations and signed Business Associate
Agreements (BAAs). Consumer-facing APIs, such as the public versions of
ChatGPT, Claude, and Gemini, do not offer BAAs and are not appropriate for use
with protected health information. Regardless of deployment model, access
controls, audit logging, and egress restrictions remain the covered entity's
responsibility. The commercial LLM landscape is evolving rapidly, and specific
model capabilities, compliance certifications, and vendor offerings described
here will continue to change. Teams should verify current status before
selecting a deployment approach.

\section{Testing}

Evaluation is integral to the SDLC. Its foundation is comparison of system
output against a reference standard, and mishandling that reference standard is
a common beginner mistake that invalidates results.

Maintaining a strict separation between training and test sets is essential. It
is easy to make apparently correct decisions on the test set using information
gleaned from it. For example, a developer might treat a term as a good predictor
because it works on the test set, or try many parameter values and keep
whichever performs best on the test set. As a rule, accuracy on genuinely new
data will be much lower than accuracy on a test set the classifier has been
tuned to. In a clean experiment, you never run on or even inspect the test set
during development. Instead, you reserve a development set for that purpose and
run a single final experiment on the untouched test set only after all
parameters are fixed. A test set that was never consulted during development is
the closest available proxy for real-world performance.

\begin{figure}[htbp]
\centering
\includegraphics[width=0.5\linewidth]{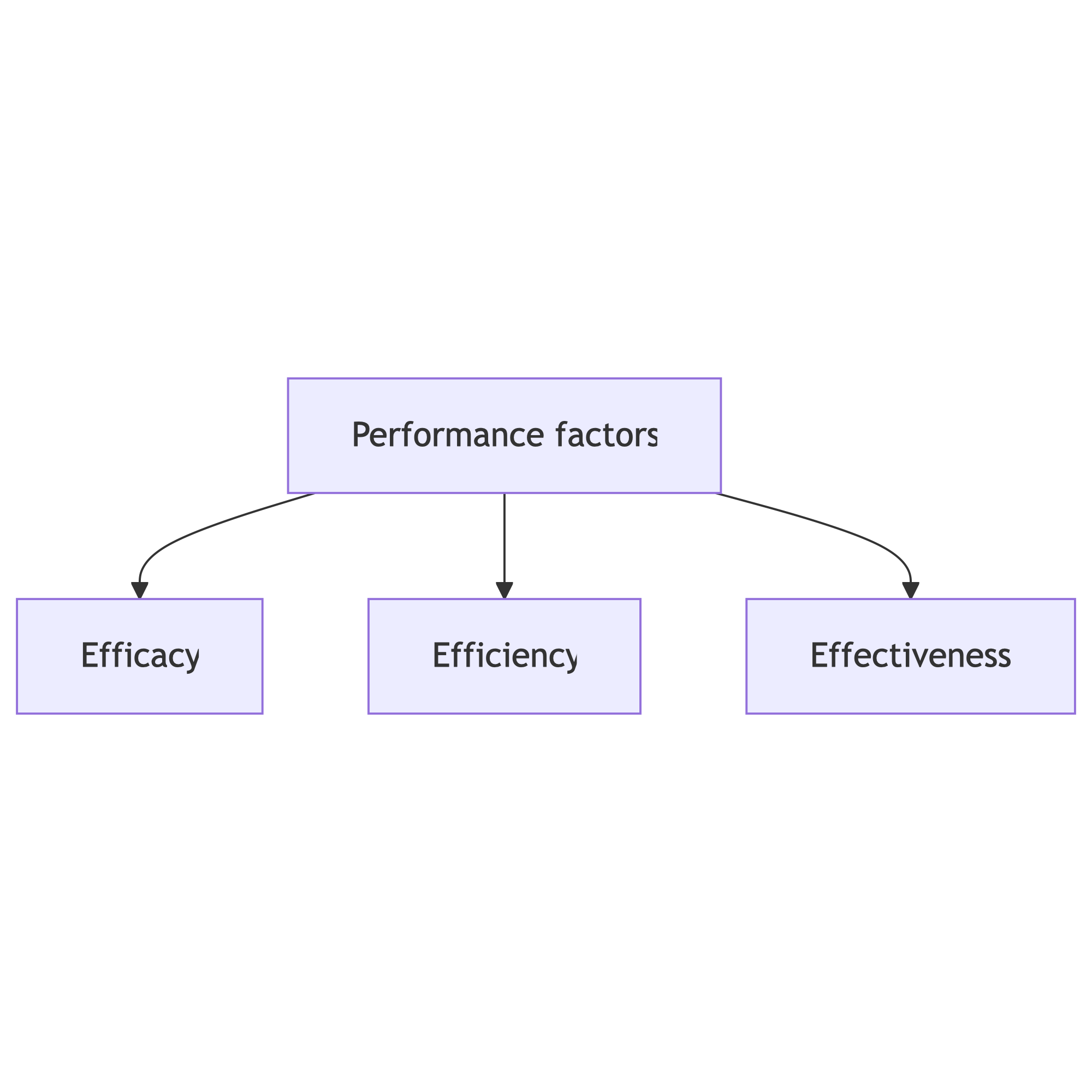}
\caption{Performance factors in system evaluation.}
\label{fig:performance}
\end{figure}

\textbf{Efficacy.} The standard measures are precision, recall, and F1.
Precision is the fraction of system-identified mentions that are correct---it
reflects how often the system is right when it makes an assertion. Recall is the
fraction of all true mentions in the text that the system identified---it
reflects how completely the system finds what is present. The two measures trade
off: a conservative system will favor precision over recall, while an aggressive
system will do the reverse. The F1 score is the harmonic mean of precision and
recall, combining them into a single value that penalizes large imbalances
between the two and facilitates comparison across systems. For LLM-based
systems, standard F1 evaluation is complicated by non-determinism and prompt
sensitivity. The same prompt can produce different outputs across runs, and small
changes in prompt wording can shift F1 by a substantial margin without any
change to the underlying model. Evaluation results are therefore tied to the
specific prompt used and should not be generalized to the model as a whole.

\textbf{Efficiency.} Efficiency measures computational performance, typically
expressed as throughput measured at the level of tokens, sentences, or
documents. For traditional NLP pipelines, throughput is predictable and scales
with hardware, making benchmarking and optimization straightforward for larger
projects. For LLM-based systems, per-document token costs and API latency
introduce economic and time constraints that must be factored into deployment
planning from the outset, and cost projections should be validated during
feasibility analysis rather than discovered at deployment.

\textbf{Effectiveness.} The system's ability to achieve its actual goal. It
usually equals efficacy, but for some use cases precision or recall matters
more, and extracting some instances with high precision can outweigh extracting
all of them. Published work often reports only efficacy, yet for an applied
system efficacy may not reflect effectiveness. If the goal is simply to
determine whether a patient has a given finding, mention-level counts are
irrelevant so long as the finding is identified at least once. Computing
effectiveness may require rolling mention-level results up to the document level.
A stricter interpretation, counting a document as a true positive only when all
its mentions are found, would make effectiveness lower than mention-level
performance. For LLM-based systems, hallucinated mentions introduce false
positives that can inflate apparent recall at the mention level while reducing
effectiveness at the document or patient level. Auditing must account for both
missed and invented mentions.

Efficiency is where the trade-off between traditional and LLM-based approaches
is most visible: a well-designed rule-based or encoder-based system can process
large clinical corpora at far greater speed and a fraction of the cost of a
comparable LLM-based pipeline. Selecting the right tool for the task matters
more than the ability to claim cutting-edge technology. When a simpler, faster
method achieves the required effectiveness even if lower efficacy, it is the
better choice.

\subsection{Error Analysis}

A formal error analysis manually examines cases where system output diverged
from annotation to find the cause. Errors fall into two kinds:

\begin{itemize}
  \item \textbf{Systematic errors} share a common cause across every instance.
  A term consistently missed because it is absent from the dictionary, or a
  sentence structure that the classifier always misreads, are systematic errors.
  Because the cause is uniform, the fix is tractable: update the dictionary,
  revise the rule, or add representative training examples for the pattern.
  Systematic errors are the most productive target for iteration because a
  single fix resolves many failures at once.

  \item \textbf{Random errors}, such as novel misspellings, highly ambiguous
  phrases, or extremely rare contexts, differ from one another and have no
  single root cause. They cannot be fixed systematically without adding so many
  special cases that precision or recall degrades elsewhere. Random errors
  represent the practical ceiling of the system for a given input domain. The
  appropriate response is to characterize them, document the limitation, and
  determine whether the residual error rate is acceptable for the use case
  rather than attempting to eliminate them entirely.
\end{itemize}

For LLM-based systems, hallucinations introduce a third category that fits
neither type. A hallucinated mention is not caused by a missing rule or an
unusual input pattern---it is generated by the model independently of what the
source text actually contains. Because hallucinations can be fluent and
contextually plausible, they are not always apparent on casual inspection and
require deliberate auditing against the source document. Error analysis for
LLM-based systems must therefore examine both false negatives (missed mentions)
and false positives (invented ones), and should test whether error patterns
correlate with prompt wording, input length, or concept type. Errors that shift
with prompt changes are a signal that the system is sensitive to surface form
rather than grounded in the clinical content.

\section{Deployment}

In software development, deployment (or \textbf{release}) is the milestone at
which the application is delivered to the customer. For research information
extraction, deployment is when the system is run on the target corpus and the
data is extracted. Several decisions must be resolved.

\textbf{Data access.} Clinical data is governed by privacy and confidentiality
regulations, so any person or process that runs the system must be authorized to
access the data. Clinical EMRs are often physically and logically separate from
the research environment. In the VA, for instance, clinicians read and write the
live EMR through VistA. To avoid accidental corruption of active records, most
retrospective studies draw from a read-only copy maintained in the Corporate
Data Warehouse (CDW), which is refreshed on a regular schedule and requires
separate authorization through Institutional Review Board (IRB) approval and a
data use agreement. Access to the CDW does not carry over from access to the
live system. This separation of live clinical records from research data is a
common governance pattern across health systems and should be confirmed as part
of feasibility planning. The research copy may also be transformed relative to
the source---mapped to standard terminologies, de-identified, or
restructured---and those transformations must be understood before building or
deploying an extraction system against it.

\textbf{Computing environment.} Depending on data volume and processing
frequency, deployment may require a scalable, distributed environment. A single
machine may suffice for a small corpus, but larger volumes benefit from
processing spread across multiple nodes. With HIPAA-compliant cloud services now
available, many organizations deploy large processing jobs to the cloud rather
than buying expensive hardware for one-time use. For LLM-based systems handling
sensitive clinical text, local deployment on organizational infrastructure is a
viable alternative that eliminates protected health information (PHI)
transmission to external services entirely.

\textbf{Repeatability.} When a system will be reused, a compiled distribution
package and clear documentation allow others to deploy and use it without the
original developer. Documentation should be written for two distinct audiences
with different needs. Developer documentation covers system deployment:
environment requirements, configuration, execution steps, and known failure
modes. Data user documentation covers working with system output and should be
grounded in the concept sheet developed during analysis: each output field
should be traceable to a defined concept, with the same definitions, scope
boundaries, and inclusion criteria the annotators used. It should also report
all performance characteristics---precision, recall, F1, and effectiveness at
the relevant level of analysis---so that downstream users understand the
expected error rate before drawing conclusions from the data. Data
transformations applied during extraction or post-processing should be described
explicitly, as should known limitations for specific use cases. Conflating these
two audiences in a single document serves neither well. A data analyst consuming
extraction results does not need to know how to configure a pipeline, and a
developer redeploying the system does not need a clinical interpretation guide.
Targeted documentation for each audience reduces misuse of output and lowers
the barrier to reuse.

\textbf{Human oversight.} LLM-based systems should not be deployed in fully
automated mode for clinical extraction without human review. The risk of
hallucinated mentions that are fluent and contextually plausible means that
automated output cannot be treated as ground truth. A structured review
step---sampling output against source documents at regular
intervals---is the minimum acceptable safeguard. Systems should also log inputs,
outputs, and model version at each run to support error tracing and revalidation
when model behavior changes.

\section{Maintenance}

An NLP system should keep performing at or above its deployment level, but
several hard-to-predict factors affect performance, chiefly unforeseen changes
in input and the fluid nature of language itself.

Language is not static. It is a dynamic, largely ergodic process
\cite{JurafskyD.2008a}. Sapir introduced the notion of unconscious linguistic drift
from his study of dialect evolution \cite{Sapir1921}. Language changes through
geographic separation and adaptation to new circumstances, including new
communication technologies \cite{Lupyan2015}, and diversification can be
morphologic (word forms), syntactic (sentence structure), or semantic (meaning).
Using word embeddings and mixed linear regression, Hamilton et al.\ \cite{Hamilton2016}
measured semantic drift in verbs and nouns via cosine similarity across corpora
stratified by decade, finding greater change in verbs globally and in nouns
locally. They concluded that shifting verb usage reflects traditional linguistic
drift, while shifting noun usage reflects cultural change, such as new words for
new technologies \cite{Gentner1988}.

The biomedical domain shows the same pattern: new terms such as ``Zika'' enter
the record, and coding systems evolve (International Classification of Diseases,
Ninth Revision, Clinical Modification [ICD-9-CM] to ICD-10-CM). Environment and
subdomain shape clinical language as well. Harris demonstrated the existence of
sublanguages mathematically \cite{Harris1991}, and Friedman et al.\ applied his
work to the clinical domain \cite{Friedman2002}, showing how semantic word
classes and specialized term meanings mark sublanguages that can be
differentiated by clinical scope \cite{Patterson2011}, specialty
\cite{Doing-Harris2013}, and disease \cite{Bernhardt2005}.

Systems used over time, or across inputs that vary in sublanguage (such as text
from different institutions), must be validated. Rule-based systems require
their rules to be re-examined and tested regularly. Machine-learning systems
must be periodically re-evaluated and likely retrained. Carrell et al.\
documented the difficulty of adapting a system across health care environments
\cite{Carrell2017}. In short, linguistic drift, new lexicon (Ebola, Zika),
evolving coding systems, changing documentation guidelines, and sublanguage
variation across subdomains all mean that rules, patterns, and models must be
updated, and systems applied over time require regular revalidation.

LLM-based systems are subject to an additional source of drift that has no
equivalent in traditional NLP: the underlying model can change without any
change to the system code. Hosted model providers update, deprecate, and replace
model versions on their own schedules, and a system validated against one
version may produce materially different output after an update. This model
version drift is harder to detect than linguistic drift because it requires no
change in the input data to manifest. Regular revalidation is therefore even
more critical for LLM-based systems than for traditional ones, and system logs
must record the exact model version used at each run so that performance changes
can be traced to their source.

\subsection{The SDLC Is a Cycle}

\begin{figure}[htbp]
\centering
\includegraphics[width=0.55\linewidth]{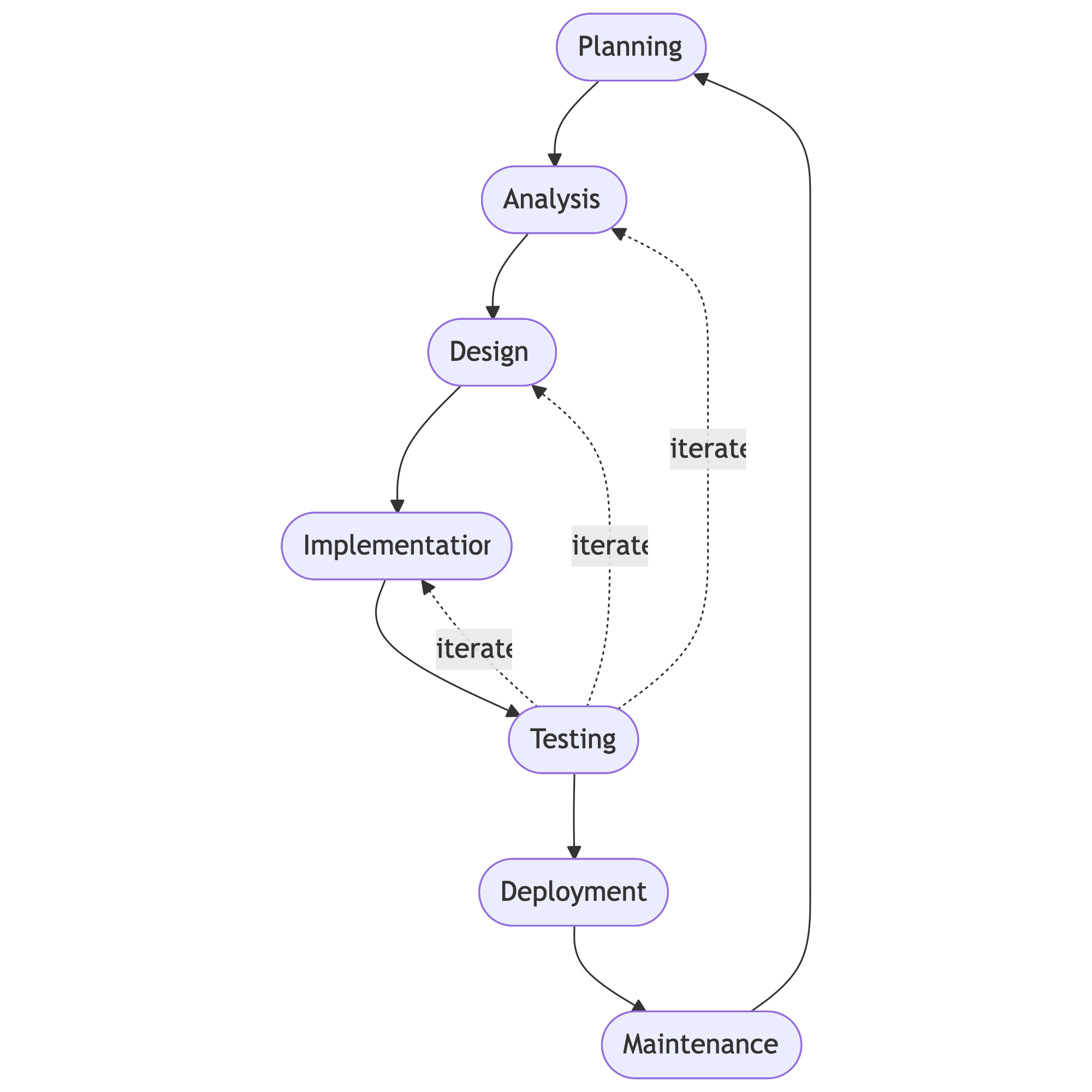}
\caption{The Systems Development Life Cycle as an iterative cycle.}
\label{fig:sdlc}
\end{figure}

Although presented sequentially, the SDLC is a cycle (Figure~\ref{fig:sdlc}).
Deployment ends one cycle. Maintenance, which itself requires planning,
analysis, design, implementation, and testing, begins the next. Depending on
what planning and analysis reveal, a new cycle may call for a major redesign or
a minor update.

Within each cycle the arrows are bidirectional. If analysis reveals that
concepts are too ambiguous or too poorly documented to extract, planning for new
concepts may be needed. Design may demand further analysis, and implementation
may force design changes.

Development also proceeds through smaller \textbf{iterations}. When error
analysis finds a systematic error, the team assesses the benefit of fixing it,
designs and implements the fix, then re-tests and re-analyzes. Iterations
continue until no more feasibly fixable systematic errors remain. Overall,
building an information extraction system is inherently iterative.

\section{Conclusions}

Regardless of a project's size, following the steps of the Systems Development
Life Cycle improves the chances of success. This paper has demonstrated how the
SDLC applies to developing clinical NLP systems, from planning and feasibility
analysis through annotation, design, implementation, testing, deployment, and
ongoing maintenance, reframing information extraction not as an algorithmic
exercise but as a process to be managed deliberately from start to finish.

The rapid expansion of large language model capabilities over the last several
years has made this argument both more timely and more urgent. A well-prompted
model can return structured output from clinical text within minutes, without
annotation, without a reference standard, and without a test set. That speed
creates the impression that the process is optional. It is not. LLMs are
genuinely capable tools that expand what is possible in clinical NLP, but they
introduce their own failure modes that the SDLC is designed to surface and
manage: hallucination, prompt sensitivity, non-determinism, model version drift,
and data governance constraints. Feasibility analysis determines whether a
concept is extractable. Annotation establishes the reference standard that makes
evaluation meaningful. Testing exposes reliability problems that a single
prototype cannot reveal. Deployment planning governs who and what can access
clinical data. Maintenance anticipates the drift that makes a validated system
degrade over time.

Quick results are not the same as accurate results, and a system that has not
been analyzed, designed, validated, and governed is not a useable system
regardless of how compelling its output appears on first inspection. The SDLC is
what converts a promising prototype into a clinical tool that performs reliably,
operates within appropriate data governance constraints, and can be trusted by
the researchers and clinicians who depend on it.

\section*{CRediT authorship contribution statement}

\textbf{Olga V. Patterson:} Conceptualization, Writing -- original draft.
\textbf{South B:} Writing -- review \& editing.
\textbf{Workman TL:} Writing -- review \& editing.
\textbf{DuVall SL:} Writing -- review \& editing.

\section*{Declaration of generative AI and AI-assisted technologies}

The original intellectual content of this manuscript was written by the authors. Editorial revisions, structural improvements, and formatting were supported by Claude (Anthropic), a large language model assistant. The authors reviewed and take full responsibility for all content.


\end{document}